\documentclass{article}
\usepackage{hyperref}
\usepackage[sort,compress]{natbib}
\usepackage{mdframed}
\usepackage{fvextra}
\usepackage{graphicx} 
\usepackage{tabularx}
\usepackage{booktabs} 
\usepackage[normalem]{ulem}
\usepackage{placeins}
\usepackage{authblk} 
\usepackage[singlelinecheck=false]{caption}

\usepackage[a4paper,          
            left=2.0cm,           
            right=2.0cm,
            top=2.0cm,
            bottom=2.0cm,
            bindingoffset=0.0cm,
            footskip=0.0cm]{geometry}

\title{Neural circuit function inference with LLMs}

\author[1,2]{Yijie Yin}
\author[2,1]{Albert Cardona}

\affil[1]{Department of Physiology, Development and Neuroscience, University of Cambridge, Cambridge, UK}
\affil[2]{MRC Laboratory of Molecular Biology, Cambridge, UK}

\begin{document}

\maketitle
\begin{abstract}
The success of connectome mapping now shifts the challenge of understanding the nervous system to the interpretation of neural circuits. 
Here, we devise a new automated method, LLantia (\textbf{LL}M \textbf{a}utomated \textbf{n}eural circui\textbf{t} \textbf{i}nference and \textbf{a}nalysis), to systematically infer neural circuit function and the role of its component neural cell types.
Our approach distills descriptions of cell type function from the literature and, in combination with the connectome, then infers the function for all other cell types, which serves as a basis for subsequent neural circuit function inference.
Results are structured hierarchically, with different possible circuit functions organised under multiple possible behavioural and physiological contexts, and each circuit function composed of subcircuit descriptions alongside relevant cell types to facilitate both backtracking to known, published information and support further experimental research.
We illustrate our method by inferring cell type function for all cell types of the adult fruit fly brain and for select broader circuits within, and validate our findings, including by cross-checking with literature published after the release date of our analysis.
\end{abstract} 

\section{Introduction}
The pace of connectome mapping is accelerating \cite{xu_enhanced_2017, hayworth_gas_2020, tavakoli_light-microscopy-based_2025, meirovitch_smartem_2026, bosch_nondestructive_2025, sievers_connectomic_2024, helmstaedter_synaptic-resolution_2026}. A purpose of mapping connectomes is to facilitate understanding of neural circuit function, in combination with molecular, functional and behavioural data. But the sheer scale of the data threatens to hinder, rather than facilitate, the formulation of models and experiments to test our understanding of how circuits work.

An approach to reduce the scale of the problem is to identify cell types by morphology and synaptic connectivity, shrinking e.g., the ca. 140,000-neuron connectome of the fruit fly \textit{Drosophila melanogaster} to the connectivity matrix of ca. 9,000 cell types \cite{dorkenwald_neuronal_2024, schlegel_whole-brain_2024, berg_sexual_2025, bates_distributed_2026-1}. 
On the basis of cell types and their specific genetic driver lines, neural circuit function can be studied by delivering a stimulus to specific cell types while monitoring the outcome in select downstream cell types; hence, focusing connectome analysis at the cell type level precisely matches the resolution of functional experiments.
Alas, connectomes such as that of the fruit fly present surprisingly high density of connectivity: within 5 hops, ca. 80\% of cell types are synaptically connected \cite{yin_connectome_2025}. Circuits relating all possible combinations between sensory neurons and descending or motor neurons are dense and recurrent, complicating the interpretation of their function, in addition to being very numerous.

The number of cell types, the dense connectivity between them, the vast array of input-output combinations of polysynaptic pathways traversing each cell type, the multimodal and combinatorial nature of sensory stimuli and motor output commands, and the abundant yet sparse literature covering only some cell types, together pose the practical problem of how to effectively explore all possible combinations while encompassing all available information in a time-effective manner.

Here, we introduce a method to combine known cell type function as extracted from the literature with circuit structure from the connectome, crafted as a prompt to an open weight LLM, to systematically infer a putative function for cell types and the circuits they are a component of.
We apply our method to all cell types of the \textit{Drosophila melanogaster} brain connectome and to a select set of brain circuits. We validate our results quantitatively and qualitatively by comparing automatically inferred descriptions of cell type and circuit function with literature published before and after the release date of the LLM and the date of application of our method to the available data.

\section{Results}

\subsection{Automated extraction of cell type function from the literature}\label{sec:lit_search}

\begin{figure}
    \centering
    \includegraphics[width=\textwidth]{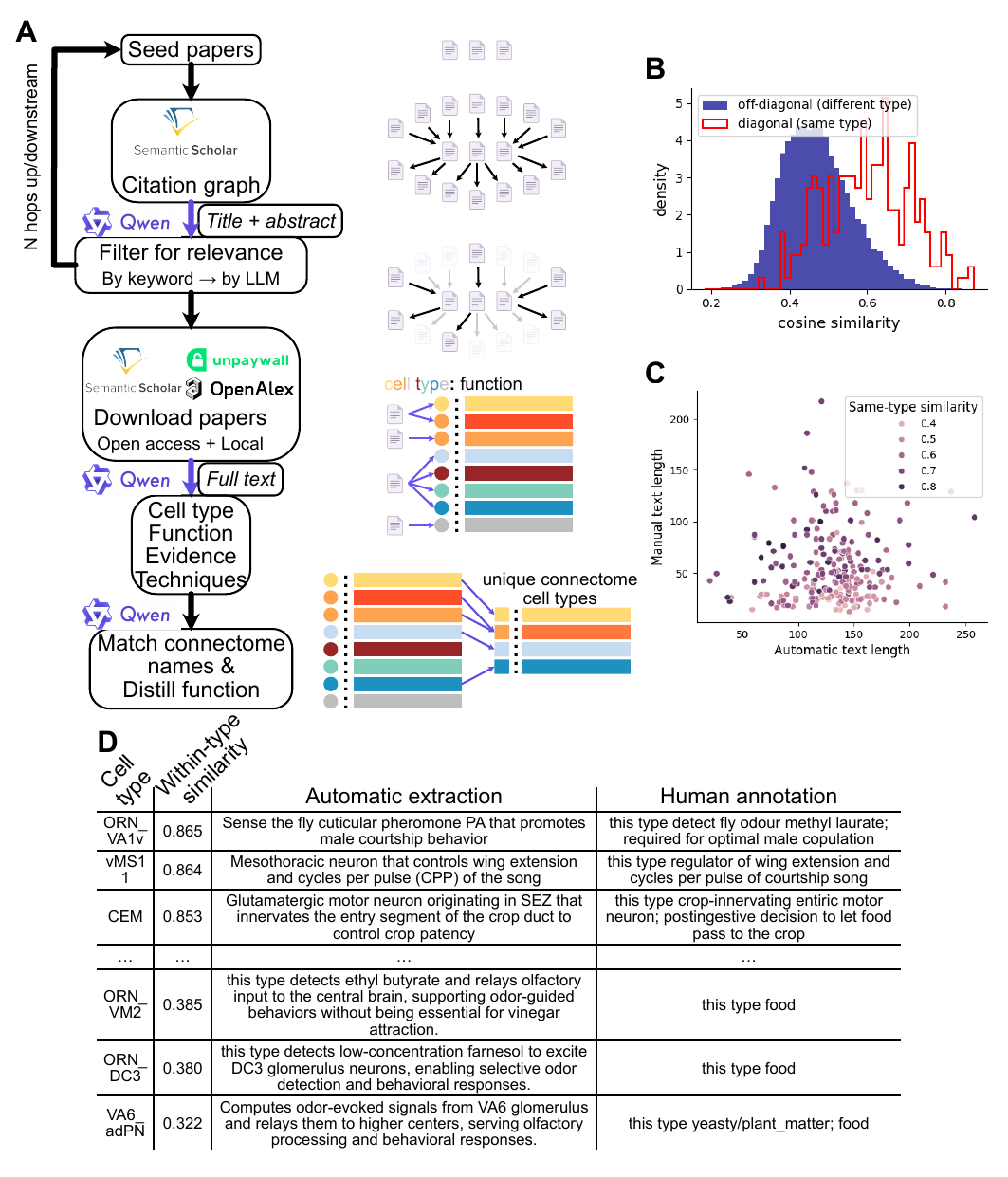}
    \caption{Extracting neuron function from the literature. \textbf{A.} Our pipeline iteratively finds papers from seed paper through citation graph, filtering for relevance at each step. The relevant papers are then downloaded (when available), and the full text is given to an LLM to extract the cell types tested and their functions. The cell types extracted from the papers are then matched to cell types in the connectome, and when a cell type is studied in multiple papers, the descriptions of the functions are summarised. \textbf{B.} The cosine similarity of the embeddings between automatically- vs. manually-extracted neuron function from the literature, between the same type and different types (Methods \ref{method:embed_dist_between_conditions}). \textbf{C.} Lengths of cell function descriptions for LLM-generated and manually-curated approaches, for cell types shared across the two collections, coloured by the embedding similarity of the descriptions. \textbf{D.} Example summaries for different within-type similarities.}
    \label{fig:lit_survey}
\end{figure}

There are $\sim9,000$ cell types in the fruit fly brain defined by neuroblast of origin, morphology and synaptic connectivity \cite{schlegel_whole-brain_2024}. Each cell type participates of unique circuits and presumably serves a different role in circuit function.
The literature on the function of cell types (i.e., the \textit{known}) is sparse (i.e, very incomplete), scattered, and constantly-evolving: both revised and expanded.

We set out to collect as comprehensive a list as possible of all cell type functions for all cell types that appear in the published literature.
Recent manual efforts to aggregate such literature \cite{bates_distributed_2026-1, yin_connectome_2025} rely on human labour and hence here we devised a strategy to automate this process.

\textit{Prior work.} There exist AI-assisted literature review tools, based on (1) keyword/semantic search (e.g., \href{https://github.com/GXL-ai/paperclip}{paperclip}, \href{https://www.scite.ai}{Scite}, \href{https://www.consensus.app}{Consensus}), which do not cover the relevant literature comprehensively, since neurons are studied in high-dimensional sensory-motor contexts and often published with neuron names which may not correspond to the cell type names as found in connectomic datasets; or (2) citation-graphs (e.g., \href{https://www.litmaps.com/}{Litmaps}, \href{https://www.researchrabbit.ai/}{ResearchRabbit}) which are available as commercial applications and do not expose application programming interfaces (API), and thus their use is either very costly or limited for comprehensive literature survey.

\subsubsection{Collecting and curating a list of relevant published papers on cell type function}

Hence, we use a free citation-graph service with an open API, Semantic Scholar \cite{lo_s2orc_2020}, to broaden the reach of the literature survey in search of descriptions of cell type function.
Starting from 114 hand-picked "seed" papers, we query the open Semantic Scholar API for upstream (papers the seed papers cite) and downstream (papers that cite seed papers) neighboring papers in the citation graph, 2 hops away. At each hop, apply relevance filters before adding the papers as seeds for the next round of citation retrieval. In particular, we assess relevance by first pattern matching for keywords in the title and abstract (such as ``\textit{Drosophila}" and ``neuron(s)") which discards many irrelevant papers, and then for the remainder we search for experiments reporting neural function in \textit{Drosophila} by feeding the titles and abstracts to a small LLM (Qwen3.5-4B).

For papers that meet the relevance criteria, we fetch open-access manuscript files (PDFs) from the internet using a combination of \href{https://www.semanticscholar.org/}{Semantic Scholar} \cite{lo_s2orc_2020}, \href{https://unpaywall.org/}{Unpaywall} \cite{piwowar_state_2018} and \href{https://openalex.org/}{OpenAlex} \cite{priem_openalex_2022}. Our pipeline is also able to utilize locally-downloaded manuscript files.

We next extract cell types and their corresponding cell type functions from the filtered and downloaded manuscripts with a specially crafted prompt for a large LLM (Qwen3.6-35B).
As expected, the cell type names returned do not always match names in the connectomics datasets, for the following reasons: 

\begin{enumerate}
    \item The cell type is referred to with a different name, i.e, a synonym: e.g., MBON-$\alpha$'1 corresponds to MBON15 \cite{li_connectome_2020};
    \item The cell type name parsed by the LLM partially contains the connectome cell type name, e.g., P9 (DNp09) corresponds to DNp09; 
    \item Many connectome cell types are studied together due to e.g., limitation in genetic access, as in, the genetic driver line used contains more than one cell type in its expression pattern. 
\end{enumerate}

We addressed the above as follows. To maximise the number of cell types extracted for further analysis, we manually screened the output to compile a list of cell type synonyms. 
Further, we crafted an LLM prompt to suggest possible options from the cell type names in the connectome based on partial string matches; the prompt requests to choose a single option where possible, and we then incorporated them into the dataset.
While genetic-driver-line-to-cell-type matching could potentially be extracted from databases such as \href{https://neuronbridge.janelia.org/}{NeuronBridge}, \href{https://www.janelia.org/project-team/flylight}{FlyLight}, and \href{https://www.virtualflybrain.org/}{VirtualFlyBrain} \cite{meissner_split-gal4_2025, court_virtual_2023, mclachlan_vfb-mcp_2026}, many papers report on experiments done on sets of neurons that include more than one cell type. In such cases, we skip these cell type names to avoid attributing the parsed function to \textit{all} cell types of that set in the manuscript.

Finally, for specific cell types that have been studied in multiple papers, we craft a prompt for an LLM (Qwen3.6-4B) to consolidate the function of the cell type across papers, to both unify the cell type function description and keep it concise (maximum 30 words).

Our pipeline implementation caches the result of iterative runs such as paper downloading and parsing. Hence, rerunning the pipeline is incremental and incurs only the costs for new papers added as further seeds or found through changes to the citation graph.

From the 114 seed papers and using 2-hops in the citation graph, our pipeline collected 20,126 papers, of which 11,819 did not pass the keyword filter, and only 3,665 were selected by the relevance filters. Of these, only 1,966 were downloaded (given academic journal paywalls that prevented downloading the rest), from which 9,766 statements on cell type function were extracted (Suppl. file \ref{supp: lit_summary_all}), of which only 2,058 matched identified cell types in fruit fly connectomic datasets, which on aggregate corresponded to 411 unique cell types.
Therefore, we have now collected the known cell type function for 411 of the ca. 9,000 cell types in the fruit fly connectome in the form of limited-length descriptive sentence for each (Suppl. file \ref{supp: lit_summary_tidy}).
The necessity of limiting the length of the descriptive sentences obeys the need to fit a number of them in further prompts to LLMs (see below).

Our approach automatically distills information from the literature on cell type function but equally applies to further properties of cell types, such as neurotransmitter signatures or receptor expression, and can be adapted to extract arbitrary information on any specified item.

\subsubsection{Evaluation of the automatically extracted cell type function descriptions}
We quantify the accuracy of the automatically collected cell type functions by comparing with the manually extracted ones, curated by expert neuroscientists \cite{bates_distributed_2026-1, yin_connectome_2025}.
We take the subset of cell types present in both the automated and the human-curated lists and, for each, embed both descriptions of its function as vectors with a text embedding model. Specifically, we use Qwen3-Embedding-8B (quantised to Q8\_0, GGUF), served locally by llama-server (llama.cpp) in embedding mode and accessed via its OpenAI-compatible /v1/embeddings endpoint, applied to a modified version of each description in which every cell type name has been replaced uniformly by "this type" to avoid biasing the comparison through name matching.
We then measure the distance between the embedding of the manually curated cell type function description sentence and the automatically extracted one.
We found that the embedding similarity is larger across manual and automated descriptions within cell types than across cell types (Figure \ref{fig:lit_survey}B) (n=240 pairs, Wilcoxon W=27798, p=1.51e-35, median paired gap=0.125, Methods \ref{method:embed_dist_between_conditions}), suggesting that the automated cell type function extraction worked as expected. Moreover, as expected, if we do not replace the cell type name with "this type", the within-type embedding similarity is higher (Figure \ref{supp: manual_vs_lit_original_name}).

Note that some cell types are known to have a similar function, e.g., in the fruit fly brain, the output neurons of the associative learning and memory centre (the Mushroom Body), even with each being uniquely identifiable morphologically and genetically, all encode valence and guide memory-based action selection \cite{aso_mushroom_2014}. Hence, that some cross-type comparisons score as high as within-type is expected in these cases. Indeed, clusters can be observed in the heatmap comparing embeddings of manually- and automatically-generated neuron function summaries (Figure \ref{supp: manual_vs_lit_heatmap}).

Further, within-cell-type embedding similarity can be affected by the following factors: 1) Neither manual nor automatic extraction of neuron function is comprehensive in coverage, so different papers might have been used by the automated vs. manual extraction for the same cell type; 2) the manual and automatic pipelines may be summarising neuron function on different levels of brevity. For instance, results from the automated inference include information on the responses to particular odours and on the behavioural consequences, whereas manual summaries can be far more succinct (Figure \ref{fig:lit_survey}D; such as when distilled from Figure 1 in \cite{zheng_structured_2022}). Therefore, quantitatively we would expect descriptions of function for the same cell type with different text lengths to score lower in the cosine similarity of their embeddings (Figure \ref{fig:lit_survey}C).

\subsection{Inferring a putative function for each cell type}
\label{sec:cell_type_function}
\begin{figure}
    \centering
    \includegraphics[width=\textwidth]{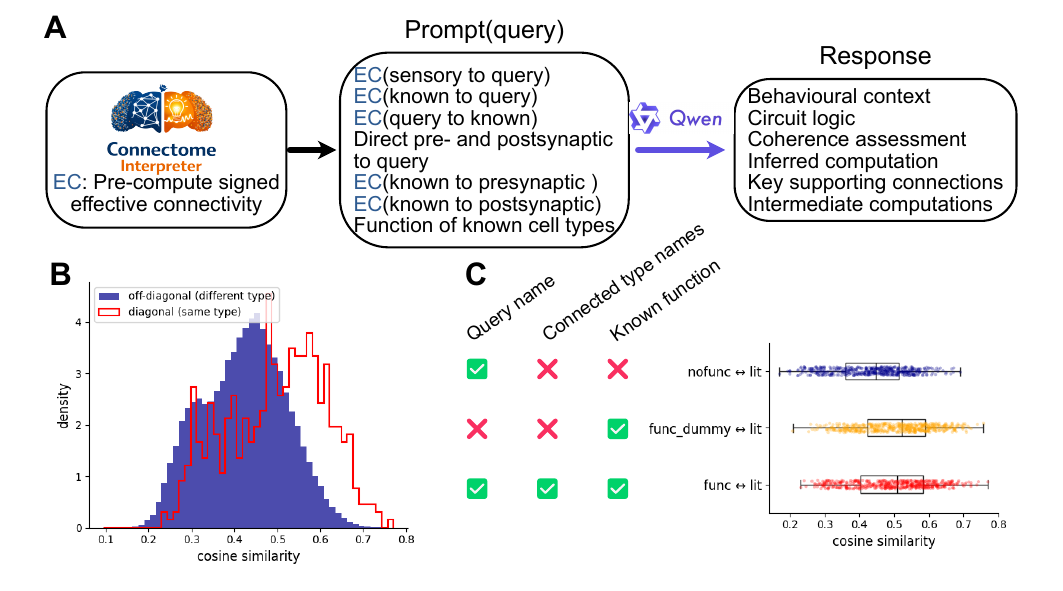}
    \caption{Inference of cell type function by combining the connectome with the literature.
    \textbf{A.} We engineer a prompt for each cell type in the fruit fly connectome, with two parameters: the connection strength threshold and the maximum polysynaptic path length. Information for the connectome is incorporated into the prompt in two ways: first, as a list of cell types directly pre- or postsynaptic to the query cell type (together with the connection weight and the sign: excitatory or inhibitory); second, as the signed effective connectivity (constrained by the connection weight threshold and the maximum path length) from (1) sensory neurons to the query cell type, (2) the cell types of known function (the "known") to the query cell type, (3) the query cell type to the known, (4) the known to the neurons directly pre- or postsynaptic to the query cell type. 
    Alongside the summarised connectome data we incorporate into the prompt a compendium of succinct text descriptions of the function of the known cell types (known because there is literature on them) that participate of the included polysynaptic pathways to/from each query cell type.
    The prompt excludes any known function of the query cell type.
    We submit the prompt to an LLM (Qwen3.6-35B) together with a request for a response structured into 6 sections: the behavioral context in which the inferred cell type function applies, the circuit logic that supports the function, a coherence assessment between the response and the information provided in the prompt, the inferred computation of the circuit the query cell type is part of, the key supporting synaptic connections to other cell types to facilitate evaluation of the response and subsequent experimental validation, and the intermediate computations in circuits upstream that support the inferred function for the cell type.
    \textbf{B.} We validate the inference of cell type function by comparing the inferred functions with the extracted functions, for the subset of 488 neuron cell types of known function in the fly connectome (both manually- and automatically-curated). By design, the known function of each cell type was excluded from the prompt. Specifically, we compare the embedding in an LLM (Qwen3-embedding:8B) of the text description of each cell type function as produced by either another LLM (Qwen3.6-35B following the procedure in panel A) or as present in the compendium of text descriptions of cell type function. There are two comparisons of embeddings: within and across cell types, i.e, across the diagonal or off-diagonal (Figure \ref{supp: hypo_vs_lit_heatmap}). 
    The distributions of diagonal values and mean-across-row off-diagonal values are statistically different (n=488 pairs, Wilcoxon W=97139, p=1.30e-33, median paired gap=0.088; see Methods \ref{method:embed_dist_between_conditions}), indicating that the description of the inferred cell type function is closer to the description of the function as extracted from the literature for the same cell type than across cell types.
    \textbf{C.} We control for the effect of the LLM memory versus its analytic capacity. We compute the cosine similarity of the embeddings of the description of cell type function between the literature-extracted function and the inferred function (see Panel A), under three conditions. First (top: nofunc $\leftrightarrow$ lit), we modify the prompt by removing the compendium of descriptions of known cell type functions, and replacing all cell type names other than the query with dummy names, to test for the effect of the LLM retaining a memory trace of the query cell type function description (from the LLM training data). The distribution of cosine similarity scores is statistically different (Wilcoxon W=13773, z=-14.7, p$<10^{-30}$, rank-biserial r=-0.769, 77\% of the scores present a lower value) than the full prompt (bottom: func $\leftrightarrow$ lit).
    Second (middle: func\_dummy $\leftrightarrow$ lit), by replacing all cell type names for dummies, including the query cell type, but still including the compendium of descriptions of known cell type functions, to test for the analytical capacity of the LLM. While the distribution of cosine similarity scores is statistically different (Wilcoxon W=46707, p=0.0000325, rank-biserial r=-0.217, 41\% of the scores present a lower value) than the full prompt (bottom: func $\leftrightarrow$ lit), the distributions are much closer than for the first control.
    }
    \label{fig:single_cell_type}
\end{figure}

Next, we set out to automatically infer the function of each cell type in the fly brain by combining the descriptions of the known functions of some cell types with the constraining architecture of the whole brain connectome.

Above, we prepared a description of the function of many cell types studied experimentally, as reported in the literature.
Here, we further prepare the connectome for its use in an LLM prompt in combination with the descriptions of cell type functions.

We pre-compute excitatory/inhibitory ('signed') effective connectivity (i.e., the multiplication of the normalized synaptic weights across consecutive edges in the wiring diagram \cite{eschbach_circuits_2020, li_connectome_2020}) from the sensory neurons and from those with known functions to the query cell type, using the Connectome Interpreter \cite{yin_connectome_2025}.
We also pre-compute the signed effective connectivity from the query cell type to the neurons with known functions.
Further, to aid in inferring the function that the target cell type performs in the local circuit, we include the synaptic connections from the direct pre- and postsynaptic neurons (with a customisable connection weight threshold), as well as the signed effective connectivity from known neurons to these pres- and postsynaptic neurons (Figure \ref{fig:single_cell_type}A).

We exclude known functions of the query cell type to prevent biasing the inference.
Such exclusion is necessary because the known function may be a subset of the actual function, given the context-dependent nature of published experiments, and the fact that for most neurons only a subset of all possible input and output pathways to the query cell type were studied experimentally.

We construct a structured prompt that is a function of the cell type to query, the threshold applied to synaptic connection weights, and the maximum polysynaptic path length to consider, and comprises all available information (Figure \ref{fig:single_cell_type}A).
We input the prompt to an LLM (Qwen3.6-35B) together with a request for a structured output response in JSON format.
The structured output lists the behavioural contexts in which the query cell type was inferred to be operating under according to the literature for the known cell types, the circuit logic in which the cell type is embedded, the coherence assessment of the hypothesis on cell type function given the information provided in the prompt, the inferred computations that the cell type participates in, key supporting synaptic connections to facilitate backtracking the inference and follow-up experimental work, and the intermediate computations performed by circuits converging onto the cell type.

We applied our analysis pipeline to the whole adult fruit fly brain, generating cell type function hypotheses for ca. 9,000 cell types (Suppl. file \ref{supp: single_neuron_hypotheses}).

Furthermore, to maximise accessibility, we prepared a \href{https://drive.google.com/file/d/1uvXI1N-qt_H6HJS8yz6TbnUwLFCaC60H/view?usp=sharing}{Google Colab notebook} which exposes multiple steps of prompt generation, in order to enable readers to experiment with custom changes to the prompt and its processing with any LLM of their choosing.

\subsubsection{Validation by comparing with literature published after the LLM release date}

To validate the approach and the results, we test the predictive power of the hypotheses, by testing the performance of predictions of neuronal cell type function that cannot have been present in the training data of the LLM, Qwen3.6, which was first \href{https://github.com/QwenLM/Qwen3.6}{released in April 2026}, and we downloaded on 13th of May 2026. Therefore we search for papers published after this date that contained reports on fruit fly neuronal cell type function.

At the time of writing, not many new papers have been released since April 2026, but we found one, for which we compare the experimental characterisation of the function of a neuron cell type against cell type functions inferred by our method.
Briefly, Coban, Harris et al. \cite{coban_olfactory_2026}, released on 21st of June, 2026, show that the neuronal cell type named LoVPN (corresponding to LTe72 in the FlyWire/FAFB connectome \cite{dorkenwald_neuronal_2024, schlegel_whole-brain_2024}) presents an elevated neuronal activity in response to visual input in the form of LED light stimulation upon removal of the antennae, which mediate other sensory modalities. Further, they detail how the cell type LHPV6m1 receives a significant amount of its input from the antennal lobe, and is directly presynaptic to LoVPN. Optogenetically silencing LHPV6m1, which is here equivalent to removing the antennal input, also results in improved visual learning.

With our method as introduced here, we inferred the following cell type function for LoVPN. Quoting from the structured response (full response available in Suppl. file \ref{supp: single_neuron_hypotheses}): 

\begin{quote}
\itshape
\textbf{Inferred computation:} ``Context-dependent visual salience detector [grounded]: The circuit selectively amplifies visual object features (vertical bars, small objects) only when competing sensory modalities (olfaction, thermosensation, motion) are below threshold, effectively gating visual attention during states of high sensory load; it acts as a memory tagging signal for visual events that are not already explained by other sensory drives, directing downstream learning centers (PAM/MBON) to encode visual associations only in neutral or exploratory contexts where other cues do not dominate behavior [likely].''

\medskip 

\textbf{Key supporting connections:} ``Strong excitatory drive from MeTu1/MeTu4/LC10 visual inputs combined with the dominant inhibitory drive from CB1318 (excited by DA3 appetitive odor and visual bar) and \textbf{LHPV6m1} (excited by thermo/hygro inputs), creating a clear multimodal gating mechanism.''
\end{quote}

The inferred cell type function for LoVPN is remarkably on point when compared to the results reported by Coban, Harris et al. 2026, down to the gating function of other sensory modalities over vision mediated precisely by neuronal cell types presynaptic to LoVPN such as LHPV6m1.

\subsubsection{Quantitative validation of inferred cell type functions}
In the first step in our method to infer cell type function, we mine the published literature to distill succinct cell type function descriptions, both manually and automatically. Here, we test whether these literature summaries match the cell type functions inferred by our method (Figure \ref{fig:single_cell_type}A) in order to assess accuracy.  
To this end, we calculate the similarity of the embeddings in an LLM (Qwen3-embedding:8B) of the two sorts of descriptions of cell type functions (as distilled from the literature vs. inferred with our method), both within and across cell types. We found that the embeddings are closer for the same cell types than across different cell types (Figure \ref{fig:single_cell_type}B; n=488 pairs, Wilcoxon W=97139, p=1.30e-33, median paired gap=0.088).

Further, LLMs have been trained with a gargantuan corpus of data that includes most if not all of the available scientific literature including preprints. Hence, we sought to disambiguate whether the inferred cell type functions are merely the product of a memory recall by the LLM, or the result of an analysis of the connectome plus known cell type functions as enclosed in the prompt. To this end, we devised two tests: one in which we control for memory recall by testing for a possible use of the cell type name as a retrieval key, and another in which we assess the effect of anonymizing all cell type names in the performance of the LLM towards inferring cell type function.

In the first test (Figure \ref{fig:single_cell_type}C, top: nofunc $ \leftrightarrow $ lit), we found that when withholding both the name and the description of the function of cell types of known function, the performance diminishes, indicating that the name of the query cell type plus the connectome as the signed effective connectivity is generally not sufficient for an LLM to yield an accurate description of a cell type function. But when a cell type is well studied (e.g., abundantly present in the literature such as the cell type T4 for visual motion detection \cite{maisak_directional_2013, gruntman_simple_2018, strother_emergence_2017, currier_infrequent_2025}) and hence better represented in the training data of the LLM, the inferred function better matches the known function. Compare them below:

\begin{quote}
\itshape
\textbf{Cell type: T4a}

\textbf{Cosine similarity:} 0.55 

\textbf{Function as distilled from the literature:} ``T4a ON front-to-back motion; orientation selective".

\textbf{Inferred computation (nofunc $ \leftrightarrow $ lit):} ``[grounded] T4a computes a directionally selective (leftward) response that is spatially sharpened and gated by global visual activity state, to distinguish local ipsilateral leftward motion from global self-generated visual flow or contralateral motion. [likely] It performs adaptive gain control of direction selectivity based on overall visual activity, preventing response saturation during high global motion (e.g. during fly rotation) so it only responds to local motion features relevant for navigation. [speculative] It provides visual feedback to steering motor circuits (via its excitatory outputs to 943, 7263, 7058) to guide leftward steering behavior in response to local visual landmarks.''
\end{quote}

Further examples of well studied neurons are listed in Suppl. file \ref{supp: single_neuron_hypotheses}. 

In contrast, for a cell type not extensively studied (i.e., without abundant reports in the literature) when the model was trained (e.g., CB0499 \cite{tastekin_connectomics_2025}), the inferred computation tends to be generic and vague, possibly applying to many neurons:

\begin{quote}
\itshape

\textbf{Cell type: CB0499}

\textbf{Cosine similarity:} 0.40 

\textbf{Function as distilled from the literature:} ``Proboscis extension (PER) hub interneuron that mediates proboscis extension reflex, contributes to sucrose feeding burst initiation."

\textbf{Inferred computation (nofunc $ \leftrightarrow $ lit):} ``[grounded] CB0499 computes a lateralized, state-dependent signal that is active only when left-hemisphere sensory or motor signals are dominant over right-hemisphere signals, and when global engagement signals (mediated by 4504) are present. [Likely] It acts as a left-preferring decision gate for sensorimotor behaviors, integrating sensory evidence from the left side with state signals to trigger or modulate left-preferring motor outputs, while being suppressed by conflicting left-side signals or contralateral dominant signals. [Speculative] It could be involved in maintaining persistent left-side spatial or attentional representations during navigation or foraging tasks, given the positive feedback loop in 9149\_left that sustains activity after initial input.''
\end{quote}

Above, note how the inferred computation does not indicate any specific sensory modality or motor response.

In the second test for isolating the effect of the analysis from that of a direct memory recall we substitute cell type names with dummy names, while likewise preserving data on signed effective connectivity and known cell type functions. Without the cell type names, we expect the inferred descriptions of cell type function to be based solely on the connectivity information and known functions of the connected partners in the prompt, rather than on a potential memory recall. The similarity of the embeddings of the dummy-name-based descriptions of neural function and those distilled from literature are \textit{on par} with those of the full-prompt (Figure \ref{fig:single_cell_type}C, compare func\_dummy $ \leftrightarrow$ lit in the middle with func $ \leftrightarrow $ lit at the bottom), indicating that, using our structured prompt, the LLM is capable of devising a cell type function description independently of what each cell type is named. In other words, the response of the LLM is not due to a mere memory recall keyed by cell type name.

\subsection{Inferring circuit function}
\label{sec:circuit_function}

\begin{figure}
    \centering
    \includegraphics[width=\textwidth]{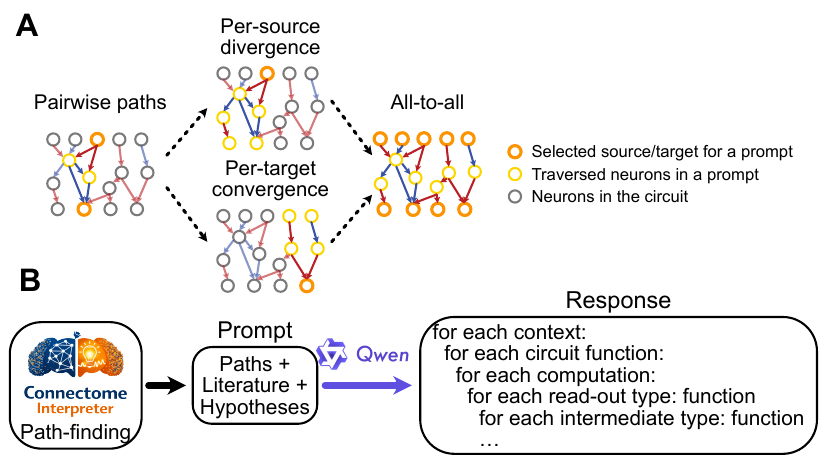}
    \caption{Generation of a description of circuit function. \textbf{A.} We feed synaptic connectivity and cell type function to the LLM hierarchically: starting with paths between a source-target pair (left, optional), then proceeding with the aggregation of per-source (mid-top) and per-target (mid-bottom) divergence and convergence, respectively, and finally combining descriptions of function from all sources to all targets (right). \textbf{B.} Each prompt includes the precomputed paths (via the Connectome Interpreter), the literature-distilled summaries of cell type function, and function descriptions for each component cell type (Section \ref{fig:single_cell_type}). The response structure is hierarchical, defining multiple neuronal cell type and circuit functions sorted into different operational contexts (behavioural and physiological).}
    \label{fig:circuit_analysis}
\end{figure}

We sought to formulate a framework for analysing neural circuits that considers as much of the structure and the components of the circuit as possible, alongside all the available information about the function of its component neuronal cell types.
In considering whole circuits rather than individual neuronal cell types, we need a strategy to express the circuit structure and known cell type function in a format that fits within the context window of an LLM (e.g., 262,144 tokens for Qwen3.6-35B, which we use hereafter) (see also Methods \ref{method:flyconn_context_length} for number of tokens required to represent a fly connectome).

We define circuits as the subgraph comprising all neuronal cell types integrating all possible polysynaptic pathways relating a defined set of input and output neuronal cell types.
We structure our approach as a series of hierarchical prompts.

First, we construct a prompt with all possible polysynaptic pathways from source to target cell types, constrained by a maximum polysynaptic path length and a minimum synaptic weight threshold.
Recurrent connections are unrolled, which has implications for the choice of the value of the maximum path length parameter.
This initial prompt includes the compendium of descriptions of known cell type functions for all included cell types, combining those extracted from the literature and the more numerous ones inferred with our method above (Section \ref{sec:cell_type_function}).
The requested structured response details a function for each input-output pair of cell types.

Second, we construct a prompt (see examples in Suppl. file \ref{supp: circuit_hypotheses}) with the prior structured response, requesting a synthesis of the description of the function of all paths from the set of input cell types that converge onto each target cell type. In parallel, we likewise construct another prompt with the symmetric request: a synthesis of the description of the function of all paths diverging from each input cell type to the corresponding set of output cell types.

Third, we collate the two responses in the prior step and craft a prompt (see the prompts in Suppl. file \ref{supp: circuit_hypotheses}) that requests a synthesis which is the description of the function of the entire circuit defined between all specified input and output cell types.

Our approach handles both small and large circuits, providing the means to comprehensively analyse more and larger circuits while considering multiple sources of information for all its component parts and retaining the necessary ability to backtrack any statements on circuit and cell type function to its sources.

The prompt templates are part of our open source software suite, LLantia (\url{https://github.com/YijieYin/llm_circuit_analysis/tree/main}).

\subsubsection{Formatting circuit structure for the LLM prompt}
We devised a textual representation of the circuit structure that would reduce redundancy as much as possible while also dispensing with the need for any graph traversal at the inference stage (i.e., in the processing of a response to the prompt).
We chose to transform the circuit into a layered representation with the Information Flow algorithm \cite{schlegel_information_2021} as implemented in the Connectome Interpreter. Hence cell types are assigned into layers represented by continuous values (i.e., a floating-point number). The layer of a cell type signifies the average number of synaptic hops from the source cell types. We sort all cell types in the circuit by their layer and list them in the prompt, with each cell type accompanied by a list of its input and output cell types.

\begin{quote}
\small
\begin{Verbatim}[breaklines,breakanywhere]
    -> type_side  [layer number, projects to type_side_1, type_side_2, ..., type_side_n] ([TARGET]) (known function (lit): ; hypothesis (prior analysis): ):
      - from type_side_1 ([SOURCE]) [layer number]: w=weight, Excit/Inhib
      - from type_side_2 ([SOURCE]) [layer number]: w=weight, Excit/Inhib
      - ...
      - from type_side_n ([SOURCE]) [layer number]: w=weight, Excit/Inhib
      (Note: content in "()" is included when applicable.)
\end{Verbatim}
\end{quote}

While downstream target cell types are listed in each entry for each cell type, more details are given on the upstream connections for each cell type, since each neuron integrates upstream information from various sources but sends the same information to all its downstream partners.

\subsubsection{Structure of the LLM response describing circuit function}
\label{sec:circuit_response_structure}

We specify a structured response to capture relevant information across the scales of neural circuit operation. Two factors motivate our approach. First, the fact that neural circuits appear hierarchical in their operation: the component cell types are structured into subcircuits (not necessarily non-overlapping) that each implements a specific function and processes a subset of the information that flows through the circuit, and the circuit under study can be defined in terms of a collection of such subcircuits. Second, the requirement for backtracking the statements in the response from the top-level overall circuit function down to subcircuits and component cell types, and further into the published literature from which cell type function was distilled.

We further structure the response into a multiplicity of responses, one for each possible behavioural or physiological context under which the chosen circuit may be operating. For example when satiated or hungry, or in a virgin versus mated state, or even simply when under multiple sensory inputs that enter into conflict. And further, under any such combination that emerges as relevant in considering the relevant cell types and their functions as distilled from the literature, since reported experiments were conducted under a specific paradigm, which may differ across published papers that study the same cell type.

Our specified response structure is thus as follows: 

\begin{quote}
\small
\begin{Verbatim}[breaklines,breakanywhere]
{{
  "contexts": {{
    "<context, in your own words>": {{
      "computations": [
        {{
          "computation": "the variable computed and the operation on it",
          "subcircuits":
            {{"<target>":
                {{
                  "function": "one sentence summary of what the target signals in this context",
                  "contributors":
                    {{
                      "<contributor 1>": [<+/->, one sentence summary of what this contributor signals in this context],
                      "<contributor 2>": [...],
                      ...
                      "<contributor n>": [<+/->, one sentence summary of what this contributor signals in this context]
                    }}
                }},
              "<target 2>": {{...}},
              ...
              "<target n>": {{...}}
            }},
          "falsification": "strongest problem found: disconfirming evidence or a self-defeating consequence, and whether it survives",
          "load_bearing": "what the circuit's behaviour loses if the computation is removed"
        }}, 
        {{
          "computation": "...",
          "subcircuits": {{...}},
          "falsification": "...",
          "load_bearing": "..."
        }}, 
        ...
      ],
      "circuit_function": "how the computations integrate into what the circuit does in this context",
      "cannot_do": "a computation not supported by the wiring, and the circuit architectural reason"
    }}
  }},
  "predictions": [
    {{"perturbation": "<cell, activate/silence>", "observable": "...", "direction": "...",
      "context": "<context(s); contrasts between contexts are most diagnostic>",
      "rests_on": ["<cell/edge> <+/->"], "falsified_if": "..."}}
  ],
  "open_questions": "...",
  "confidence": "high/medium/low: coherence, completeness, consistency with the connectivity matrix"
}}
\end{Verbatim}
\end{quote}

\subsubsection{Application to circuits in the fruit fly brain}
We applied our pipeline to 6 broad circuits in the fruit fly brain connectome, obtaining the following automated summaries of their function:
\begin{enumerate}
    \item \underline{Lateral horn to central complex:} (3 contexts, listing context 1 only)
    \begin{quote}
        \textbf{context:} Hungry, non-ovipositioning female fly navigating in a threat-free environment when encountering a salient food-associated odor (e.g. vinegar, ethanol, CO2) \\
        \textbf{function:} The circuit integrates food odor valence, foraging priority, and state conflict signals to generate a unified appetitive navigation output that drives targeted upwind tracking toward food sources while blocking irrelevant competing behaviors
    \end{quote}
    
    \item \underline{Mushroom body output neurons (MBONs) to central complex}: (3 contexts, listing context 1 only)
    \begin{quote}
        \textbf{context:} Hunger-driven foraging, no competing reproductive or social priorities, encountering appetitive vs aversive odor cues \\
        \textbf{function:} This circuit filters appetitive and aversive odor signals against internal hunger state to drive appropriate foraging navigation, suppress maladaptive steering, and prioritize approach to rewarding odors while avoiding aversive cues.
    \end{quote}

    \item \underline{Mushroom body output neurons (MBONs) to descending neurons (DNs)}: (3 contexts, listing context 1 only)
    \begin{quote}
        \textbf{context:} Hungry, unthreatened foraging state (no imminent looming threat, no courtship partners present, low aversive odor load) \\
        \textbf{function:} The circuit computes a unified net foraging approach drive that coordinates forward locomotion, lateralized steering toward food objects, and controlled landing to enable efficient foraging for nutrient rewards while suppressing competing behaviors that would interfere with foraging
    \end{quote}

    \item \underline{Olfactory projection neurons (PNs) to dopaminergic neurons (DANs) of the mushroom body}: (2 contexts, listing context 1 only)
    \begin{quote}
        \textbf{context:} Hungry, unmated male fly in humid, warm conditions with no rival male, receptive female, or high-priority aversive cues (geosmin, parasitic wasp pheromone) present \\
        \textbf{function:} The circuit computes safe, context-appropriate food odor signals, suppresses inappropriate foraging or avoidance responses, and drives localized upwind approach and long-term memory consolidation for unconfounded food sources when no higher-priority social or survival needs are active
    \end{quote}
    
    \item \uline{Neurons with significant effective connectivity to multiple of the 22 behaviour-based DN clusters defined in Bates et al. \cite{bates_distributed_2026-1}, used as sources, and analysed with respect to their downstream DNs of known function}: (3 contexts, listing context 1 only) 
    \begin{quote}
        \textbf{context:} Mated female low-threat foraging/oviposition state (no looming threat, no active courtship/escape, hunger or oviposition readiness present, CB4242 active) \\
        \textbf{function:} The circuit computes graded, low-priority approach and oviposition motor commands to guide the mated female toward food and oviposition substrates in the right visual field, prioritizing these behaviors only when no higher-priority threat or competing behavior is active
    \end{quote}
    
    \item \underline{From turning descending neurons to each other}: (3 contexts, listing context 1 only)
    \begin{quote}
        \textbf{context:} Slow forward walking while foraging small, non-threatening food items (e.g. rotting fruit) in the left visual field \\
        \textbf{function:} Coordinate smooth, goal-directed leftward approach to non-threatening food items during slow forward walking, while suppressing all conflicting high-stakes motor responses to prevent collision or loss of approach trajectory
    \end{quote}
\end{enumerate}

We illustrate how we prepared the analysis of each of these 6 examples by walking through the first example, on the fruit fly brain circuit from the Lateral Horn Neurons (LHNs, known for innate olfactory memory) to the Central Complex (CX, known to be responsible for navigation in the fly \cite{hulse_connectome_2021, seelig_neural_2015, green_neural_2017, turner-evans_angular_2017, matheson_neural_2022}). We begin by selecting the cell types to be included in the analyses: LHNs that have strong connections with CX neurons. We compute the effective connectivity from LHNs to all the CX neurons within 3 synaptic hops, thresholding input-normalised direct connection strength at 1\%, and selecting LHN-CX cell type pairs where the effective connectivity exceeds 0.01. This criteria returned 42 LHNs, and 93 CX neuron cell types, plus further cell types mediating polysynaptic pathways between them, structuring a complex network encompassing 1,047 cell types that is approachable with our method.

For this circuit our prompt elicits from the LLM a response listing 3 contexts, each context containing 2 computations, each computation listing 1 subcircuit, and each subcircuit highlighting a small handful of neurons. We illustrate the nature of the response below by focusing on a single element at each level, i.e., one context, one computation within that context and one cell type involved in the computation:

\begin{quote}
\itshape
\textbf{Context:} ``Hungry, non-ovipositioning female fly navigating in a threat-free environment when encountering a salient food-associated odor (e.g. vinegar, ethanol, CO2)"

\textbf{Computation:} ``Compute context-gated heading error correction signal that only triggers steering adjustments when the fly is off-course relative to the food odor source, suppressing redundant corrections when already aligned"

\textbf{FB2F\_b:} ``function": ``Corrective guidance signal for odor-guided locomotion that triggers discrete heading realignments when off-course relative to food odor",

\textbf{Contributors:} ``LHPD2a4\_a": [``+",
                  ``Signals heading deviation from appetitive odor source combined with learned reward valence"], 
\end{quote}

The complete set of contexts, computations, and neural functions present in the LLM responses for all the example circuits analysed are listed in the Supplemental file \ref{supp: circuit_hypotheses}.

\subsubsection{Backtracking the circuit analysis to the source cell types}

\begin{figure}
    \centering
    \includegraphics[width=\textwidth]{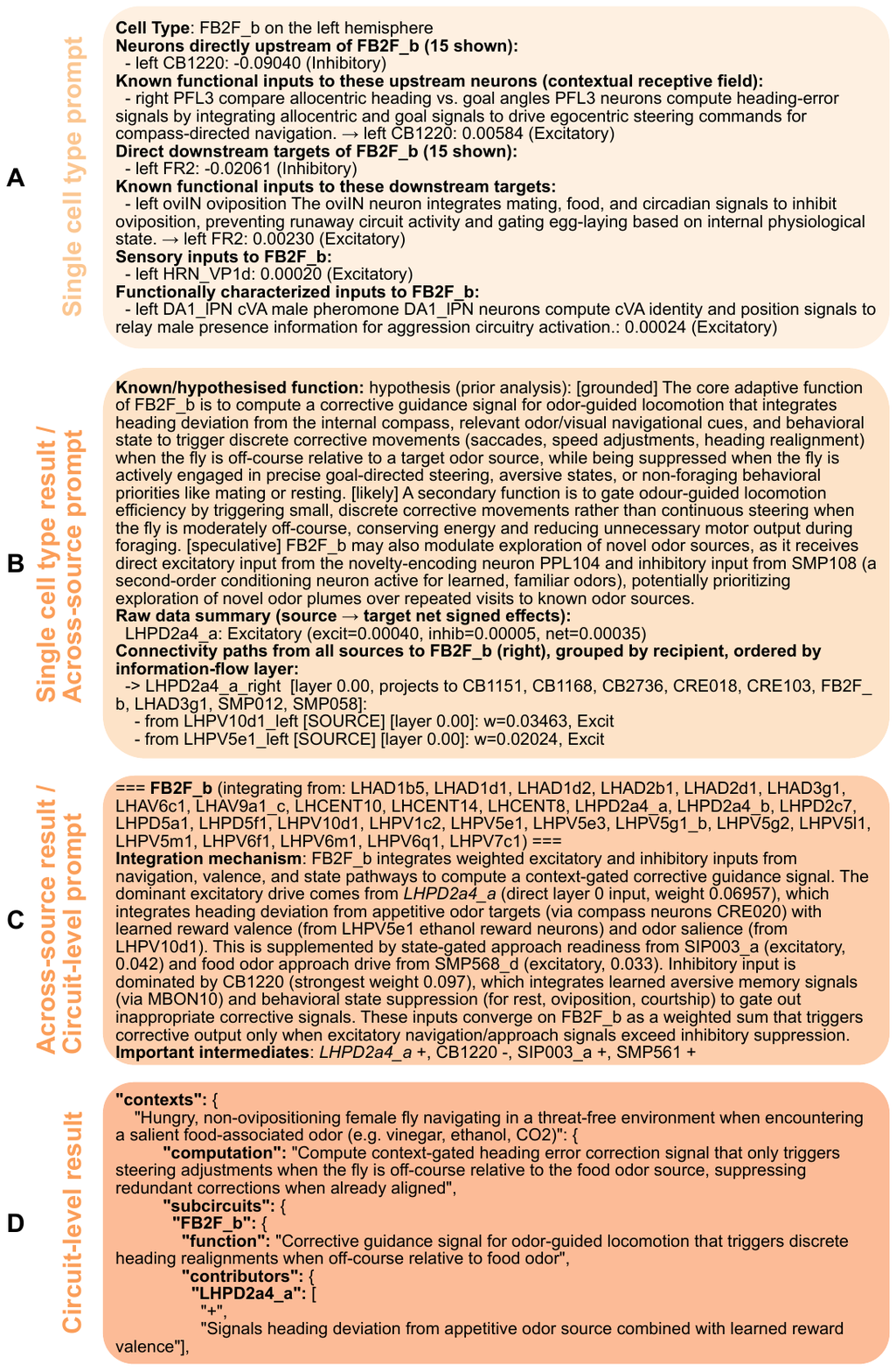}
    \caption{Sequence of prompts (showing as cropped) and their results. Each response to a prompt serves, in turn, as part of the prompt for the next query to the LLM until obtaining the response describing the circuit function.}
    \label{fig:backtrack}
\end{figure}

We illustrate the ability to backtrack across steps of the analysis for the LH to CX example illustrated above.

We demonstrate in Figure \ref{fig:backtrack} the capability of our pipeline to facilitate backtracking from the result of circuit analysis (Section \ref{sec:circuit_response_structure}) to the automatically-generated prompt for single-cell-type analysis (Section \ref{sec:cell_type_function}), so that the evidence supporting the claims can be examined.

Briefly, for the claim that the LH to CX circuit does compute a "context-gated heading error correction signal" in the context of "hungry, non-ovipositioning female fly navigating in a threat-free environment", we track back to the effecting cell types.
Starting from the excitatory effective connection from LHPD2a4\_a to FB2F\_b contained in the circuit-level result (Figure \ref{fig:backtrack}D), we examine the corresponding prompt (Figure \ref{fig:backtrack}C), which contains the result from the previous step describing the across-source integrating function of FB2F\_b, where the direct synaptic connection from LHPD2a4\_a is included in the subheading titled \textbf{integration mechanism}. Next, in examining the prompt of the across-source analysis (Figure \ref{fig:backtrack}B), we notice it includes the synaptic connectivity information for LHPD2a4\_a, which was automatically constructed via subcircuit extraction (through path-finding) and information flow. To further backtrack from the \textbf{known/hypothesised function} in the across-source prompt, we then examine the synaptic connectivity information present in the prompt (Figure \ref{fig:backtrack}A) which was used for describing the function of FB2F\_b in Section \ref{sec:cell_type_function}.

\subsubsection{Common themes across responses describing circuit function}

We observed the following common themes in the descriptions of circuit functions generated for all the examined circuits above.

First, cross-inhibition between behaviours. For instance the cell type LPLC2, known to detect looming stimulus (rapidly-approaching visual stimulus) \cite{klapoetke_ultra-selective_2017, klapoetke_functionally_2022, turner_visual_2022} effectively inhibits pMP2, known to drive courtship song \cite{lillvis_nested_2024} and DNa01, known to drive ipsiversive steering \cite{rayshubskiy_neural_2025, yang_fine-grained_2024, chen_imaging_2018}. Such cross-inhibition corresponds to the exclusionary selectivity of most behaviours, in that the animal can only execute one behaviour at a time.

Second, AND gates, as in, meeting multiple requirements simultaneously. For instance, here are the multiple requirements that the cell type FB8F\_a, which is hypothesised to "drive upwind navigation to protein-rich food sources only when oviposition priority is active and no conflicting behaviors are present", need to meet in the result of the circuit analysis. First, it has to be excited by MBON23 which "encodes context-specific food odor attraction, suppresses response to aversive or conflicting odors". Second, it has to be not inhibited by MBON25 which "encodes approach suppression, suppresses FB8F\_a when foraging is non-adaptive". And third, not inhibited by SLP024\_right which "encodes courtship priority conflict, suppresses FB8F\_a when courtship is active".

\subsubsection{Further ways to apply our framework for circuit analysis}

We have illustrated the application of LLantia, our LLM-based circuit analysis toolkit, with a bottom-up exploratory approach to inferring circuit function. For hypothesis-driven circuit analysis, the prompts can be modified to instruct the LLM to focus on the specific hypotheses and contexts. For selecting relevant cell types to limit which ones are to be included in the analysis, our methods can further compare the embedding similarity of the question with single-cell-type descriptions of cell type function generated from Section \ref{sec:cell_type_function}.

\section{Discussion}
We developed a framework, LLantia, for the automated systematic neural circuit function inference capable of handling whole connectomes plus the available literature.
We released LLantia as open-source software, readily applicable to published connectomes, supported by our Connectome Interpreter open-source software and the open-weight LLMs, and published alongside documentation for its practical application using the fruit fly connectome as a ready example.
We illustrated the application of our approach to inferring cell type function and circuit function by generating inferences for all cell types of the adult fruit fly brain and a select set of broad circuits within.
We designed LLantia to yield inferences that support backtracking to both validate the results and to directly support further experimentation by specifically relating individual cell types to the functions of the broader circuits they are a component of.

\subsection{Advantages of requiring strictly structured responses}

In addition to highly structured prompts, our framework yields specifically structured responses that serve a dual purpose.

First, the response structure makes explicit the levels of analysis requested. For example, when generating hypotheses on the function of single cell types, the response includes subfields such as \textit{key supporting connections} and \textit{coherence assessment}, directing the LLM to provide evidence for the statements in the inferred computation.

Second, a structured response in JSON format facilitates easy programmatic access and manipulation of the results in scalable analysis pipelines for downstream steps. For example, the literature search pipeline generates dictionary-like mappings from cell types to their functions, separate from the specific experimental evidence that justified the statement about the cell function. We incorporate this dictionary directly into downstream prompts.

\subsection{Modularity in pipeline design}

Connectome-based neural circuit analysis typically involves the following components: defining the question, literature search to identify the known neurons, circuit slicing from the broader whole-brain connectome, and integrating all information to achieve a conclusion on what the circuit does.

Consequently, we designed the LLantia software as a set of interactive modules, each performing one single task and designed to run independently or as sequential steps in a pipeline. The first step is the selection of a specific question in the form of a circuit for which to infer a function, alongside the necessary parameters (polysynaptic path lengths and synaptic weight thresholds) for restricting the circuit that is to be sliced out of the connectome with the Connectome Interpreter. The role of the LLM is limited to performing function inference from a deliberately crafted prompt within a limited context length (number of total tokens an LLM can process in one session). 

Modularity in design separates literature search, neural cell type function inference, and circuit function inference. Such separation supports selective execution, repurposing of components, and rerunning with different parameters or a different choice of circuit, in addition to ad-hoc adjustments of intermediate outputs (e.g., to include unpublished data). Our documentation and open source nature of our software suite renders each of these interventions straightforward.

\subsection{Value-added of LLMs for circuit function inference} 

The application of our software suite yields a description of circuit function in natural language, and in the course of its application we noted the following:

\begin{enumerate}
    \item \textbf{Inference}: By construction and execution, we demonstrate here that LLMs can indeed perform useful inference on neuron/circuit function based on natural language descriptions of synaptic connectivity and distilled information from the relevant literature.
    
    \item \textbf{Selection, abstraction, and compression}: in translating from high-dimensional circuit function inference to lower-dimensional rationales of information flow, LLMs usefully select subcircuits, and abstract and compress combinations of cell types into explanations of multisensory integration or behavioural decision making. 
    
    \item \textbf{Context-dependent inference}: Neural circuits, without changing a single synapse, can operate differently depending upon sensory and motor context \cite{taisz_generating_2023, steinfath_neural_2025}, so any account of circuit function must consider its specific context of application.
    Behavioural and physiological context is high-dimensional and multi-scale, and is described with natural language in the published literature; the conclusions drawn from each study are likewise described in natural language. LLMs can therefore operate on these descriptions, inferring, when guided through a multi-scale set of nested responses as we do, what a given neuronal cell type or synaptic connection among cell types conveys within a particular context rather than in an underdetermined landscape of possibilities. Context, by constraining the possible outputs, seems to further contribute to producing useful, specific, actionable inferences on circuit and neuron function, pushing circuit function inference beyond what mere structured synaptic connectivity could express.

    \item \textbf{Capacity} Although context (in the LLM prompt processing sense) length remains a constraint, we showed that modern, open-weight LLMs can accommodate a fairly extensive account of a neuron's synaptic connectivity, the known functions of its synaptic partners, and the subsequent inference about the circuit function. Going forward, further technical progress towards longer contexts alongside faster, cheaper GPU inference will only extend the reach of our hierarchical, modular and parallelisable framework, supporting the systematic application of our circuit function inference method to large numbers of neurons and subcircuits in ever larger connectomes, as long as sufficient information about the function of their component cell types exists.
\end{enumerate}

\subsection{Limitations}

Our framework depends on prior research reporting on the function of at least some of the neuronal cell types, which serves as the basis for inference. 

Furthermore, while intermediate and final responses on circuit and cell type function infernece are highly structured, the answers in the corresponding subfields are after all free text. We have here evaluated the responses by calculating the similarity in embedding space to descriptions extracted from the literature, for cell type function. However, we have not checked systematically whether the similarity of the embeddings corresponds to matching descriptions on neural function, rather than generally describing function in a particular context.

In addition, experimental tests of neural function typically focus on a subset of the connectivity of specific cell type of interest, whereas LLMs are explicitly directed to consider the complete synaptic connectivity into account when inferring function. More detailed validation, in which automated approaches or human experts evaluate functional agreement directly, is therefore needed, which is a major motivation for the inclusion of relevant cell types in the explanations of circuit function.

\section{Acknowledgments}

We thank the MRC LMB for core funding, the Wellcome Trust Investigator award (205038/A/16/Z) to A.C. and the ERC Synergy (project 101167460, "CircuitEvolution") to A.C.. We thank the LMB Scientific Computing team and the High Performance Cluster for computational resources. We thank Marta Costa, Aljoscha Nern, Jakob Macke, Judith Hoeller, Stephan Saafeld, Katherine Nagel for insightful discussions.
Y.Y. discloses use of Claude (Anthropic) for iterating the development of the LLantia software.

\section{Materials and Methods}
\subsection{Connectomics data}
We use the connectivity information between each neuron and neuron metadata in both FAFB/FlyWire \cite{dorkenwald_neuronal_2024, schlegel_whole-brain_2024} (v783, using the synapse predictions in \cite{yu_new_2025}) and maleCNS \cite{berg_sexual_2025} (v0.9) \textit{Drosophila melanogaster} connectomes, preprocessed \href{https://github.com/YijieYin/connectome_data_prep/tree/main}{here}, where the connection strength between two neurons is input-normalised and is a floating point number between 0 and 1. Neurotransmitter expression (here used to determine the excitatory/inhibitory sign of the connections) is based on predictions generated in \cite{eckstein_neurotransmitter_2024}.

\subsection{Compute and time consumption}
\begin{table}[ht]
\centering
\caption{Computational resources and typical wall-clock times for each pipeline
component, measured from SLURM accounting (\texttt{sacct}) on the LMB HPC
cluster. The \texttt{ml} partition runs on A100-40\,GB\,SXM4 nodes (AMD EPYC
7452, 128\,HT cores, 1\,TB RAM); the \texttt{agpu} partition runs on
RTX\,4090-24\,GB nodes (Threadripper PRO 7975WX, 64\,HT cores, 512\,GB RAM).
Quoted times are \emph{per array task} (i.e.\ per GPU-worker). The large range
for Circuit Step\,1 reflects the variation in the number of cell-type pairs
across different circuit modules.}
\label{tab:compute}
\footnotesize
\setlength{\tabcolsep}{3pt}
\begin{tabular}{p{4cm} p{5cm} l l r p{2.2cm}}
\toprule
\textbf{Step} & \textbf{Script} & \textbf{Partition} & \textbf{GPU} & \textbf{Array} & \textbf{Wall time} \\
\midrule
Paper discovery              & \texttt{paper\_discovery\_hpc.sh}              & \texttt{ml}   & 1$\times$ A100\,40G & ---  & 45\,min -- 2.5\,h \\
PDF matching \& download     & \texttt{match\_and\_download\_pdfs.sh}         & CPU           & ---                 & ---  & $\sim$5\,min \\
Function extraction          & \texttt{function\_extraction\_from\_paper.sh}  & \texttt{ml}   & 1$\times$ A100\,40G & 8    & 4 -- 5\,h \\
CT name normalisation        & \texttt{post\_extraction.sh}                   & \texttt{agpu} & 1$\times$ RTX\,4090 & ---  & 15 -- 50\,min \\
CT function summarisation    & \texttt{summarize\_cell\_functions.sh}         & \texttt{agpu} & 1$\times$ RTX\,4090 & 8    & $\sim$1\,min \\
Embedding                    & \texttt{embed\_descriptions.sh}               & \texttt{agpu} & 1$\times$ RTX\,4090 & ---  & 1 -- 6\,min \\
Neuron interpretation        & \texttt{neuron\_interpretation.sh}            & \texttt{ml}   & 1$\times$ A100\,40G & 8    & 1 -- 2\,days \\
Circuit Step\,1 (pairwise)   & \texttt{run\_step1.sh}                        & \texttt{ml}   & 1$\times$ A100\,40G & 8    & 10\,min -- 6\,h \\
Circuit Step\,2 (per target) & \texttt{run\_step2.sh}                        & \texttt{ml}   & 1$\times$ A100\,40G & 8$^a$& 5 -- 40\,min \\
Circuit Step\,2$'$ (per src) & \texttt{run\_step2\_per\_source.sh}           & \texttt{ml}   & 1$\times$ A100\,40G & 8$^a$& 5 -- 25\,min \\
Circuit Step\,3 (synthesis)  & \texttt{run\_step3.sh}                        & \texttt{ml}   & 1$\times$ A100\,40G & ---  & 2 -- 15\,min \\
\bottomrule
\multicolumn{6}{p{0.95\linewidth}}{%
  $^a$ Steps\,2 and\,2$'$ run concurrently with a \texttt{\%4} cap each,
  sharing the per-user 8-GPU limit.
  All \texttt{ml}-partition steps use Qwen3.6-35B-A3B-Anko Q8\_0;
  \texttt{agpu}-partition steps use Qwen3.5-4B Q6\_K (summarisation/normalisation)
  or Qwen3-Embedding-8B Q8\_0 (embedding).
  Context windows: neuron interpretation 60\,k; circuit steps\,1--2 up to 98\,k;
  step\,3 200\,k (with q8 KV-cache quantisation).}
\end{tabular}
\end{table}

\subsection{Counting number of tokens} 
Knowing the number of tokens a prompt consumes is useful for guiding the iterative process of prompt engineering. 
For automatically-constructed prompts, where possible, we provide example notebooks (such as \href{https://colab.research.google.com/drive/1dD9wuClQ0TJe8uDaH-iE3j59bWxjKLuC?usp=sharing}{this one}) to inspect the prompts and count the number of tokens consumed by the prompt (using the transformers Python package \cite{wolf_transformers_2020}).

\subsubsection{Counting number of tokens for an adult fly connectome}
\label{method:flyconn_context_length}
To be efficient in context length, we group connections by cell type, remove all connections $<$ 1\% of the recipient cell type's total input, and round the connection strength to 2 significant digits. Further, assuming Dale's law that each cell type releases one fast-acting neurotransmitter \cite{dale_pharmacology_1935, eckstein_neurotransmitter_2024}, and that the effect of this neurotransmitter is the same for all post-synaptic partners, we avoid repeating presynaptic cell types and the outgoing sign, by representing the connectivity information in the following format: 
\begin{quote}
\begin{Verbatim}[breaklines,breakanywhere]
    (PLP191,PLP192)a +:
        PLP161 0.01
        CL151 0.02
        CB3937 0.01
        PLP209 0.02
        ...
\end{Verbatim}
\end{quote}

As demonstrated in \href{https://colab.research.google.com/drive/1dD9wuClQ0TJe8uDaH-iE3j59bWxjKLuC?usp=sharing}{this Colab Notebook}, such lean representation takes $>$ 2 million tokens, far larger than the context window of e.g. Qwen3.6. 

\subsection{Quantifying similarity of embeddings between conditions}
\label{method:embed_dist_between_conditions}
To compare the similarity of the embeddings between two conditions (e.g. automated vs. manual literature summary) between the same vs. different cell types, we compare: 
\begin{enumerate}
    \item similarity of embeddings between the two statements on the query cell type, with 
    \item average similarity between the statements of query cell type with the statements of all cell types other than the query. 
\end{enumerate}
for all cell types. 

We thus have one pair of data points for each cell type. We hypothesise that the embedding similarity between statements on the same type should be higher than the embedding similarity between statements on different types. We thus choose Wilcoxon signed-rank test.

\section{Supplementary figures}
\renewcommand\thefigure{S\arabic{figure}}
\setcounter{figure}{0}
\begin{figure}[htbp]
    \centering
    \includegraphics[width=\textwidth]{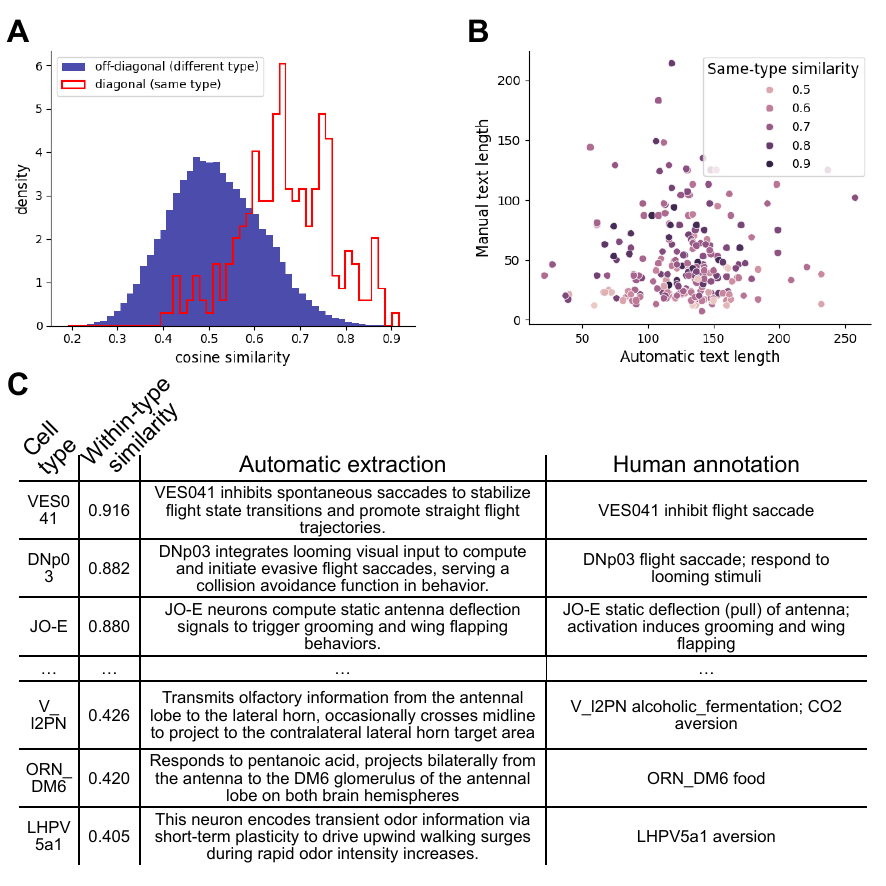}
    \caption{Comparison of manually- vs. automatically-extracted descriptions, without substituting the cell type name by "this type". \textbf{A.} The cosine similarity of the embeddings between automatically- vs. manually-extracted neuron function from the literature, between the same type and different types (Methods \ref{method:embed_dist_between_conditions}). \textbf{B.} Lengths of cell function descriptions for LLM-generated and manually-curated approaches, for cell types shared across the two collections, coloured by the embedding similarity of the descriptions. \textbf{C.} Example summaries for different within-type similarities.}.
    \label{supp: manual_vs_lit_original_name}
\end{figure}

\renewcommand\thefigure{S\arabic{figure}}
\begin{figure}[h]
    \centering
    \includegraphics[width=\textwidth]{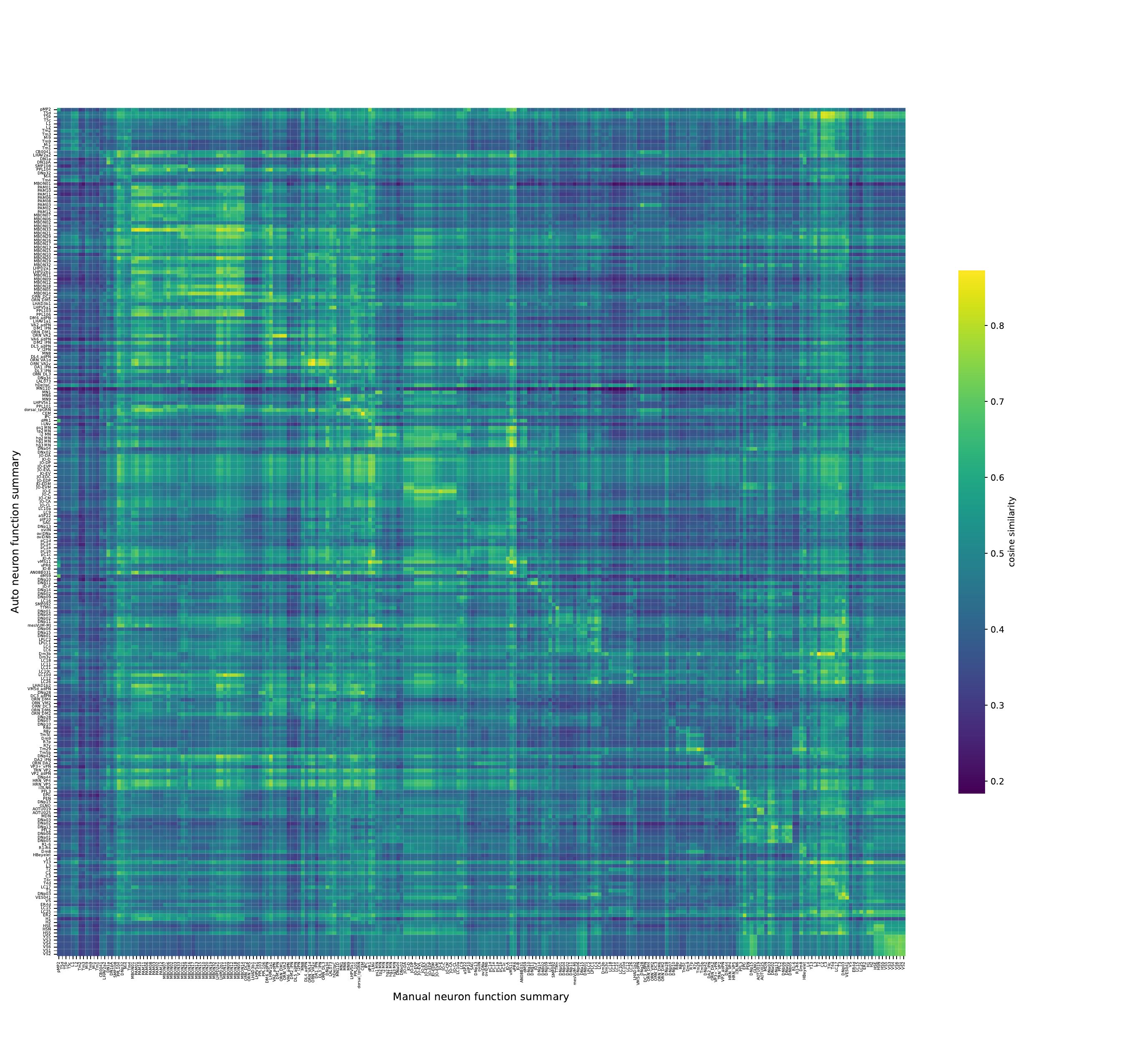}
    \caption{Embedding similarity for descriptions of cell type function between human curation and automatic extraction.}
    \label{supp: manual_vs_lit_heatmap}
\end{figure}

\renewcommand\thefigure{S\arabic{figure}}
\begin{figure}[h]
    \centering
    \includegraphics[width=\textwidth]{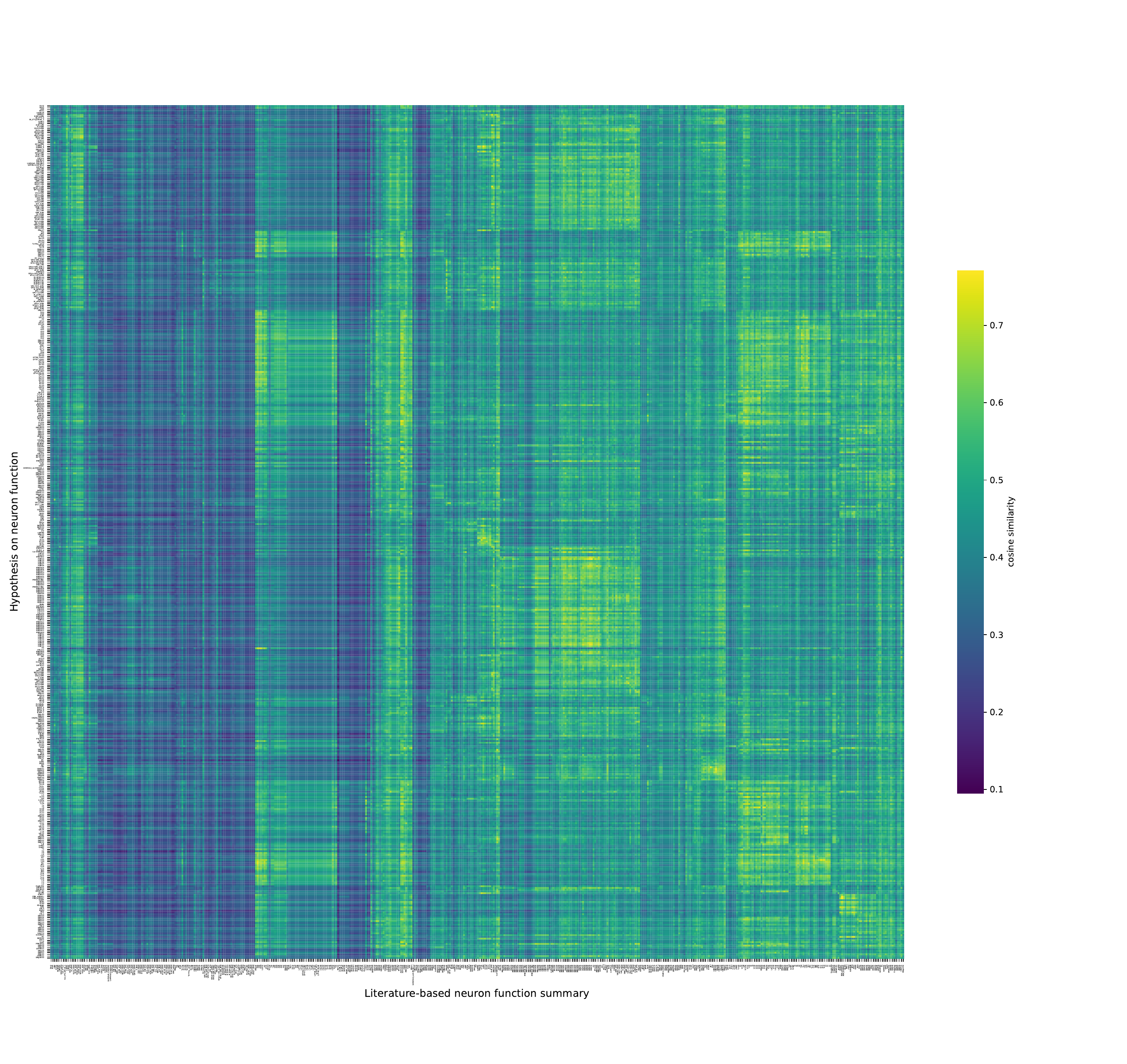}
    \caption{Embedding similarity for descriptions of cell type function between literature extraction and automatically-generated hypotheses (Section \ref{sec:cell_type_function}).}
    \label{supp: hypo_vs_lit_heatmap}
\end{figure}

\FloatBarrier
\section{Code availability}
The code and prompts used are open-source under MIT license: \url{https://github.com/YijieYin/llm_circuit_analysis/tree/main}. 

\section{Supplementary data}
\renewcommand{\theHtable}{supp.\arabic{table}}
\begin{enumerate}
    \item \label{supp: lit_summary_all} Supplementary file 1. Statements on cell type functions extracted from the literature, with partial matches to cell types in the published connectomes. 
    \item \label{supp: lit_summary_tidy} Supplementary file 2. Summarised statements on cell type functions extracted from the literature, for cell types in the published connectomes. 
    \item \label{supp: single_neuron_hypotheses} Supplementary file 3. Single-cell-type-level hypotheses for the FAFB/FlyWire connectome with/without including known functions of connected neurons, with/without converting cell types to dummy names. There are two collections: one generated on 20/June/2026 before Coban, Harris et al. 2026, and another on 17/Jul/2026 used for subsequent analyses. Deposited on Zenodo: \url{https://doi.org/10.5281/zenodo.21628534}.
    \item \label{supp: circuit_hypotheses} Supplementary file 4. A .zip file containing prompts and hypotheses generated for a number of local circuits. Deposited on Zenodo: \url{https://doi.org/10.5281/zenodo.21628534}
\end{enumerate}

\section{Software used}
\begin{table}[h]
\caption{Software used in the LLM-assisted connectome analysis pipeline.}
\label{tab:software}
\small
\begin{tabularx}{\linewidth}{llp{4.8cm}>{\raggedright\arraybackslash}X}
\hline
\textbf{Software} & \textbf{Version} & \textbf{Purpose} & \textbf{Reference} \\
\hline
Python & 3.12 & Programming language & \url{https://www.python.org} \\
\hline
\multicolumn{4}{l}{\textit{LLM inference}} \\
\multicolumn{4}{l}{\hspace{1em}\textit{Option A — local (llama.cpp + Qwen)}} \\
llama.cpp & b8346 & Local LLM server (OpenAI-compatible API) & \cite{gerganov_ggml-orgllamacpp_2026} \\
Qwen3-Embedding-8B (Q8\_0 GGUF) & --- & Text embeddings of cell type function descriptions & \url{https://huggingface.co/Qwen/Qwen3-Embedding-8B-GGUF} \\
Qwen3.6-35B-A3B-Anko (Q8\_0) & --- & Function extraction, neuron interpretation, circuit analysis & \url{https://huggingface.co/allura-org/Qwen3.6-35B-A3B-Anko} \\
Qwen3.5-4B (Q6\_K) & --- & Paper relevance filtering, cell-function summarisation & \url{https://qwen.ai/apiplatform} \\
\multicolumn{4}{l}{\hspace{1em}\textit{Option B — OpenAI API}} \\
GPT & --- & Cloud LLM (annotation \& synthesis) & \url{https://developers.openai.com/api/docs/models} \\
\multicolumn{4}{l}{\hspace{1em}\textit{Option C — Anthropic API}} \\
Claude & --- & Cloud LLM (annotation \& synthesis) & \url{https://platform.claude.com/docs/en/about-claude/pricing} \\
\multicolumn{4}{l}{\hspace{1em}\textit{API clients}} \\
openai (Python SDK) & 2.26.0 & OpenAI-compatible API client (Options A \& B) & \url{https://github.com/openai/openai-python} \\
anthropic (Python SDK) & 0.84.0 & Anthropic API client (Option C) & \url{https://github.com/anthropic/anthropic-sdk-python} \\
\hline
\multicolumn{4}{l}{\textit{Connectome analysis}} \\
connectome\_interpreter & 2.10.0 & Connectome path finding \& effective connectivity & \url{https://github.com/YijieYin/connectome_interpreter} \\
\hline
\multicolumn{4}{l}{\textit{Data processing}} \\
pandas & 2.3.3 & Tabular data handling & \cite{the_pandas_development_team_pandas-devpandas_2025} \\
NumPy & 2.2.6 & Numerical arrays & \cite{harris_array_2020} \\
\hline
\multicolumn{4}{l}{\textit{PDF processing \& matching}} \\
PyMuPDF & 1.27.2 & PDF metadata extraction & \url{https://pymupdf.readthedocs.io} \\
RapidFuzz & 3.14.5 & Fuzzy title matching & \url{https://github.com/maxbachmann/RapidFuzz} \\
\hline
\multicolumn{4}{l}{\textit{Literature discovery APIs}} \\
Semantic Scholar & --- & Citation-graph BFS \& open-access URLs & \cite{lo_s2orc_2020} \\
Unpaywall & --- & Open-access PDF retrieval & \cite{piwowar_state_2018} \\
OpenAlex & --- & Supplementary PDF locations & \cite{priem_openalex_2022} \\
\hline
\multicolumn{4}{l}{\textit{HTTP utilities}} \\
requests & 2.32.5 & HTTP downloads & \url{https://requests.readthedocs.io} \\
requests-cache & 1.3.2 & API response caching & \url{https://requests-cache.readthedocs.io} \\
\hline
\end{tabularx}
\end{table}

\clearpage
\bibliographystyle{unsrt} 

\bibliography{references}
\end{document}